# Dikkat Katmanlı Derin Sinir Ağları ile Anahtar Kare Tespiti

# Key Frame Extraction with Attention Based Deep Neural Networks

Samed Arslan, Senem Tanberk
samedarslan90@gmail.com, 0000-0003-1668-0365
Research and Innovation
Huawei Turkey Research and Development Center, Istanbul

*Özetçe*

Videolardan otomatik anahtar kare tespiti, uzun videolar için içeriği en iyi özetleyebilecek sahnelerin seçilmesi için yapılan bir çalışmadır. Videonun özetini sağlamak hızlı göz atmayı ve içerik özetlemeyi kolaylaştırmak için önemli bir görevdir. Elde edilen fotoğraflar farklı sektörlerde otomatikleştirilmiş işler (örneğin; güvenlik görüntülerinin özetlenmesi, müzik kliplerinde kullanılan farklı sahnelerin tespit edilmesi) için kullanılmaktadır. Bunun dışında ileri makine öğrenim yöntemlerinde yüksek hacimli videoların işlenmesi, ayrıca kaynak maliyeti yaratır. Elde edilen anahtar kareler; kullanılacak yöntemlere ve modellere girdi özellik olarak kullanılabilmektedir. Bu çalışmada; dikkat katmanına sahip derin bir oto kodlayıcı model kullanarak, anahtar kare tespiti için, derin öğrenmeye dayalı bir yaklaşım öneriyoruz. Önerilen yöntem, önce otomatik kodlayıcının kodlayıcı kısmını kullanarak video karelerinden öznitelikleri çıkarır ve bu nitelikler ile benzer kareleri bir arada gruplamak için k-ortalama kümeleme algoritmasını kullanarak segmentasyon uygular. Ardından, kümelerin merkezine en yakın kareler seçilerek her kümeden anahtar kareler seçilir. Metot TVSUM video veri setinde değerlendirildi ve 0,77'lik bir sınıflandırma doğruluğu elde edildi, bu da birçok mevcut metottan daha yüksek bir başarım oranını göstermektedir. Önerilen yöntem, video analizinde anahtar kare çıkarma için umut verici bir çözüm sunmaktadır ve video özetleme ve video alma gibi çeşitli uygulamalara uygulanabilir.

*Anahtar Kelimeler — Otomatik Anahtar Kare Tespiti, Video Özetleme, Derin öğrenme, Dikkat Katmanlı Öğrenme, Oto Kodlayıcılar*

*Abstract*

Automatic keyframe detection from videos is an exercise in selecting scenes that can best summarize the content for long videos. Providing a summary of the video is an important task to facilitate quick browsing and content summarization. The resulting photos are used for automated works (e.g. summarizing security footage, detecting different scenes used in music clips) in different industries. In addition, processing high-volume videos in advanced machine learning methods also creates resource costs. Keyframes obtained; It can be used as an input feature to the methods and models to be used. In this study; We propose a deep learning-based approach for keyframe detection using a deep auto-encoder model with an attention layer. The proposed method first extracts the features from the video frames using the encoder part of the autoencoder and applies segmentation using the k-means clustering algorithm to group these features and similar frames together. Then, keyframes are selected from each cluster by selecting the frames closest to the center of the clusters. The method was evaluated on the TVSUM video dataset and achieved a classification accuracy of 0.77, indicating a higher success rate than many existing methods. The proposed method offers a promising solution for key frame extraction in video analysis and can be applied to various applications such as video summarization and video retrieval.

## I. GİRİŞ

Video analizi, gözetim, eğlence ve eğitim gibi çok çeşitli alanlardaki uygulamalarla, son yıllarda giderek daha önemli bir araştırma alanı haline geldi. Anahtar kare çıkarma, video analizinde çok önemli bir adımdır, çünkü önemli olayları veya eylemleri yakalayan en bilgilendirici kareleri seçerek bir videonun içeriğini özetlemeye yardımcı olur[1]. Diğer bir deyişle, video anahtar kare çıkarmanın temel amacı, video içeriğini temsil edecek video karelerini tespit etmektir.

Tespit edilecek özet karelerin temsil özellikleri çalışma alanlarına göre farklılık gösterebilmektedir. Ancak en genel kullanım senaryosu, video akışı içerisindeki farklı sahne ve çevre unsurlarının tespit edilmesi veya genel akıştaki nesne hareket hızlarından daha farklı hızda hareket eden nesnelerin tespit edilebildiği karelerin bulunmasıdır.

Geleneksel anahtar kare çıkartmaya yönelik yöntemler genellikle, buluşsal yöntemlere veya belirli video içeriğine büyük ölçüde bağlı olabilen ve diğer videolarda kullanılmak üzere genellenemeyebilen el yapımı uygulamalara dayanır. Bu çalışmada, dikkat katmanına sahip derin bir otomatik kodlayıcı kullanarak anahtar kare çıkarma için yeni bir yaklaşım öneriyoruz. Kullanılacak model, anahtar kare çıkarma için ihtiyaç duyulan video içeriğinin özelliklerini otomatik olarak öğrenebilen bir yapıda olacaktır[2].

Önerilen yöntem, önce otomatik kodlayıcının kodlayıcı kısmını ve dikkat katmanını kullanarak video karelerinden öznitelikleri çıkarır ve daha sonra bu öznitelikleri K-means kümelemesi kullanarak kümeler. Anahtar kareler daha sonra, küme merkezine olan yakınlıklarına göre her bir kümeden en yakın olan seçilir. Otomatik kodlayıcının kodlayıcı kısmında dikkat katmanının kullanılması, video karelerindeki en belirgin özelliklerin vurgulanmasına yardımcı olur ve bu, anahtar kare çıkarma işleminin doğruluğunu artırır[3][4].

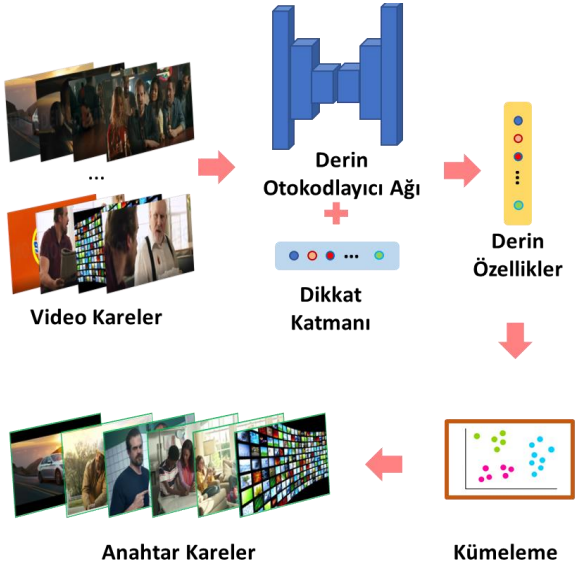

Şekil. 1. Uygulama akışı

Önerilen yöntemin etkinliğini değerlendirmek için, anahtar kare çıkarımı için bir kıyaslama veri kümesi üzerinde deneyler yapıldı. Deneysel sonuçlar, önerilen yöntemin, anahtar kare çıkarma için mevcut yöntemlerle karşılaştırılabilir olduğunu ve iyi performans gösterdiğini, mevcut yöntemlere alternatif olarak kullanılabileceğini, video özetleme ve eylem tanıma gibi çeşitli alanlarda potansiyel uygulamalara sahip olduğunu göstermektedir.

Bu teknik incelemenin geri kalanı şu şekilde düzenlenmiştir: Bölüm iki, dikkat katmanına sahip derin bir otomatik kodlayıcı kullanarak anahtar kare çıkarma için önerilen yöntemin ayrıntılı bir açıklamasını sağlar. Kullanılan veriyi detaylıca anlatır. Bölüm üç, deneysel kurulumu ve değerlendirmemizin sonuçlarını sunar. Son bölümde, çalışma sonuçları değerlendirilir ve yorumlanır.

## II. YÖNTEM

### A. Ön İşleme

Girdi olarak kullanılmak üzere, görüntü verisini makine öğrenimi modeli için düzenlemek gerekecektir. Bu adımlarda hem standardizasyon hem de normalizasyon çalışması yapılacaktır. Her adımın çalışma dökümü aşağıda verilmiştir:

İl olarak, görüntünün renk uzayı BGR'den HSV'ye dönüştürülmesidir. Bu, renk bilgisini parlaklık bilgisinden ayırmak için bilgisayar görüşünde kullanılan yaygın bir tekniktir[5].

İkincisi, görüntüyü 64 x 64 piksel boyutlarında bir kare şeklinde yeniden boyutlandırılır. Görüntüyü daha küçük bir boyuta yeniden boyutlandırmak, görüntünün en önemli görsel özelliklerini korurken model üzerindeki hesaplama yükünün azaltılmasına yardımcı olabilir. Bu adımda yeterli kaynak sağlandığı durumlarda, daha büyük çözünürlükler genellikle daha başarılı sonuçlar üretecektir.

Üçüncü olarak, görüntünün piksel değerleri 0 ila 255 aralığında normalleştirilir. Bu, tüm piksel değerlerinin tutarlı bir aralıkta olmasını sağlamak için görüntü işlemede kullanılan yaygın bir tekniktir.

Son olarak, önceden işlenmiş görüntü bir listeye eklenir. Kullanılmaya hazır hale getirilir.

Kısaca, bu adımlar bir görüntü almak, renk alanını HSV'ye dönüştürmek, daha küçük bir boyuta yeniden boyutlandırmak, piksel değerlerini normalleştirmek ve onu bir makine öğrenimi modeline beslenebilecek önceden işlenmiş kareler listesine eklemek şeklindedir.

### B. Metodlar

#### B.1. Derin Öğrenme ve Otokodlayıcılar

"Derin öğrenme", çok katmanlı sinir ağlarına dayanan makine öğrenimi algoritmalarının bir alt kümesini ifade eder. Bu algoritmalar, doğrusal olmayan dönüşümlerin birden çok katmanı aracılığıyla girdileri yinelemeli olarak dönüştürerek verilerin karmaşık temsillerini öğrenebilir[6].

"Otokodlayıcılar", veri sıkıştırma ve özellik çıkarma gibi denetimsiz öğrenme görevleri için derin öğrenmede kullanılan bir tür sinir ağı mimarisidir. Otomatik kodlayıcılar, orijinal girdi ile yeniden oluşturulmuş çıktı arasındaki farkı en aza indirerek girdi verilerini yeniden yapılandırmak için eğitilmiş bir kodlayıcı ve bir kod çözücü ağından oluşur[7].

Kodlayıcı ağı bir girdi alır ve onu, girdinin en önemli özelliklerini yakalayan daha düşük boyutlu bir temsile veya gizli alana eşler. Kod çözücü ağı daha sonra gizli temsili alır ve onu orijinal girdi alanına geri eşler[8].

Otomatik kodlayıcılar, göreve bağlı olarak farklı kayıp işlevleriyle eğitilebilir. Örneğin, veri sıkıştırmada, kayıp fonksiyonu, orijinal girdi ile yeniden oluşturulmuş çıktı arasındaki ortalama karesel hata olabilir. Özellik çıkarmada, kayıp fonksiyonu, kodlayıcı ağının girişi ve çıkışı arasındaki fark olabilir[9].

Otomatik kodlayıcıların ilginç bir özelliği, veri sıkıştırma, gürültü giderme ve anormallik algılama gibi görevler için yararlı olabilecek, verilerin kompakt ve verimli temsillerini öğrenebilmeleridir. Otomatik kodlayıcılar, önceden eğitilmiş kodlayıcı ağının bir aşağı akış görevi için özellik çıkarıcı olarak kullanıldığı aktarım öğrenimi için de kullanılabilir.

Literatürde, otomatik kodlayıcılar görüntü tanıma, konuşma tanıma ve doğal dil işleme gibi çeşitli alanlarda; özellik çıkarma veya veri sıkıştırma için bir araç olarak kullanılabilir. Araştırmacılar, model tarafından öğrenilen gizli alanı keşfetmek ve öğrenilen özelliklerin yapısını analiz etmek için otomatik kodlayıcıları da kullanabilir. Yeniden üretilebilirliği ve diğer yöntemlerle karşılaştırmayı kolaylaştırmak için otomatik kodlayıcı için kullanılan mimarinin, eğitim prosedürünün ve değerlendirme ölçümlerinin ayrıntılı açıklamalarını sağlamak önemlidir.

B.2. *Derin Öğrenme Dikkat Katmanları(Attention Layer)*

Dikkat katmanları, görüntü ve video oluşturma, metin özetleme ve makine çevirisi gibi görevlerde performanslarını artırmak için otomatik kodlayıcılara eklenebilen bir tür mekanizmadır[10].

Dikkat katmanları, modelin çıktıyla alaka düzeyine bağlı olarak girdi verilerinin farklı bölümlerine seçici olarak odaklanmasını sağlar. Bu, çıktı için önemine bağlı olarak her girdi öğesine bir ağırlık atayarak ve bu ağırlıkları girdi öğelerinin ağırlıklı bir toplamını hesaplamak için kullanarak yapılır[11].

Otomatik kodlayıcılarda, girdideki önemli özellikleri seçerek, vurgulayarak yeniden oluşturulan çıktının kalitesini artırmak için dikkat katmanları kullanılabilir. Örneğin, görüntü oluşturmada, istenen çıktıya bağlı olarak girdi görüntüsünün farklı bölgelerine seçici olarak odaklanmak için dikkat katmanları kullanılabilir. Metin özetlemede, özetin uzunluğuna ve içeriğine bağlı olarak girdi belgesindeki önemli cümlelere veya anahtar sözcüklere seçici olarak odaklanmak için dikkat katmanları kullanılabilir[12][13].

Otomatik kodlayıcılarda kullanılabilecek, küresel dikkat, yerel dikkat ve kişisel dikkat gibi farklı türde dikkat katmanları vardır. Küresel dikkat, tüm girdi dizisi için tek bir ağırlık seti hesaplarken, yerel dikkat, girdi dizisinin farklı bölümleri için farklı ağırlık kümeleri hesaplar. Çok başlı dikkat olarak da bilinen kendi kendine dikkat, birden fazla ağırlık setini paralel olarak hesaplayarak modelin girdi verilerinin farklı yönlerini yakalamasına olanak tanır.

Akademik çalışmalarda, otomatik kodlayıcılardaki dikkat katmanları, modelin görüntü ve video oluşturma, metin özetleme ve makine çevirisi gibi çeşitli görevlerdeki performansını artırmak için kullanılabilir. Araştırmacılar, farklı türdeki dikkat mekanizmalarını keşfedebilir ve doğruluk, şaşkınlık veya F1 puanı gibi ölçümleri kullanarak bunların etkililiğini değerlendirebilir[14].

B.3. *Kümeleme Yöntemleri*

Kümeleme algoritmaları, veri noktalarını benzerliklerine göre gruplandırmak için kullanılan gözetimsiz makine öğrenimi algoritmalarının bir sınıfıdır. Kümelemenin amacı; grup etiketleri hakkında herhangi bir ön bilgi olmaksızın, verilerde doğal gruplamalar veya kümeler bulmaktır[15].

Hiyerarşik kümeleme, bölüm tabanlı kümeleme, yoğunluk tabanlı kümeleme ve model tabanlı kümeleme dahil olmak üzere birkaç tür kümeleme algoritması vardır. Bu algoritmaların her biri, veri noktalarını gruplandırmak için farklı bir yaklaşım kullanır.

Hiyerarşik kümelemede, veri noktaları, mesafelerine veya benzerliklerine göre yinelemeli olarak kümeler halinde birleştirilir. Bu, istenen sayıda kümeyi elde etmek için belirli bir yükseklikte kesilebilen dendrogram adı verilen ağaç benzeri bir yapıyla sonuçlanır[16].

K-means gibi bölüm tabanlı kümeleme algoritmaları, noktalar ile bir küme merkezleri kümesi arasındaki mesafelere bağlı olarak veri noktalarını sabit sayıda kümeye böler. Merkezler, yakınsamaya ulaşılana kadar yinelemeli olarak güncellenir ve bu da verilerin kümelere bölünmesiyle sonuçlanır[17].

DBSCAN gibi yoğunluğa dayalı kümeleme algoritmaları, veri alanının yoğun bölgelerinde bulunan veri noktalarını daha düşük yoğunluklu alanlarla ayırarak gruplandırır. Bu, verilerin yoğunluğuna bağlı olarak değişen şekil ve boyutlarda kümelerle sonuçlanır.

Gauss karışım modelleri (GMM) gibi model tabanlı kümeleme algoritmaları, veri noktalarının olasılık dağılımlarının bir karışımından üretildiğini varsayar ve dağılımların parametrelerini ve verilerdeki küme sayısını tahmin etmek için istatistiksel yöntemler kullanır.

Görüntü analizi, metin madenciliği ve sosyal ağ analizi gibi çeşitli alanlardaki verileri analiz etmek için kümeleme algoritmaları kullanılabilir. Araştırmacılar, farklı türde kümeleme algoritmalarını keşfedebilir ve siluet katsayısı, saflık veya entropi gibi metrikleri kullanarak bunların etkililiğini değerlendirebilir. Girdi verileri, mesafe veya benzerlik ölçüsü ve algoritmanın performansını optimize etmek için kullanılan herhangi bir hiperparametre veya ayarlama yöntemi dahil olmak üzere, kullanılan kümeleme algoritmasının net bir tanımını sağlamak önemlidir[18][19].

B.4. *Uzaklık Metrikleri*

Mesafe ölçümleri, iki veri noktası arasındaki benzerliği veya farklılığı ölçmek için kullanılan matematiksel işlevlerdir. Makine öğreniminde, mesafe ölçümleri genellikle özellik vektörleri veya veri noktaları arasındaki mesafeyi ölçmek için kullanılır ve genellikle kümeleme, sınıflandırma ve regresyon görevlerinde kullanılır[20][21].

Öklid mesafesi, Manhattan mesafesi, kosinüs mesafesi ve Mahalanobis mesafesi dahil olmak üzere çeşitli mesafe ölçütleri vardır. Bu ölçümlerin her biri, iki veri noktası arasındaki mesafeyi hesaplamak için farklı bir formül kullanır.

Öklid mesafesi belki de en iyi bilinen mesafe metriğidir ve iki veri noktasında karşılık gelen özellik değerleri arasındaki farkların karelerinin toplamının karekökü olarak hesaplanır. Taksi mesafesi olarak da bilinen Manhattan mesafesi, karşılık gelen özellik değerleri arasındaki mutlak farkların toplamı olarak hesaplanır[22].

Kosinüs mesafesi, iki vektör arasındaki açının kosinüsünü ölçer ve genellikle iki belge veya metin parçacığı arasındaki benzerliği ölçmek için doğal dil işleme görevlerinde kullanılır[23]. Mahalanobis mesafesi, verilerdeki özellik değerlerinin kovaryansını hesaba katar ve genellikle özellikler yüksek oranda ilişkili olduğunda kullanılır.

Makine öğrenimi modellerinin performansını değerlendirmek, farklı modellerin veya algoritmaların performansını karşılaştırmak için genellikle mesafe ölçümleri kullanılır. Araştırmacılar, farklı türde mesafe ölçümlerini keşfedebilir ve doğruluk, F1 puanı veya ROC eğrisi altındaki alan gibi ölçümleri kullanarak bunların etkinliğini değerlendirebilir. Matematiksel formül ve metriğin varsayımları veya sınırlamaları da dahil olmak üzere, kullanılan mesafe metriğinin net bir tanımını sağlamak önemlidir[24].

*C. Veri Kümesi ve İçerik*

TVSum (TV Özeti), video özetleme araştırması için halka açık bir veri kümesidir. Video özetleme algoritmalarını değerlendirmek için akademik makalelerde yaygın olarak kullanılır. Video özetleme, videonun en önemli içeriğini yakalayan anahtar kareleri veya çekimleri seçerek uzun bir videonun özlü ve temsili bir özetini otomatik olarak oluşturma görevidir[25].

TVSum veri seti, haberler, talk şovlar, spor, belgeseller ve eğlence şovları dahil olmak üzere çeşitli türleri kapsayan 50 popüler TV şovundan video klipler içerir. Veri kümesi, video özetleme algoritmalarını farklı bağlamlarda değerlendirmek için uygun hale getiren çeşitli görüntüler içerir. TVSum'daki her video klip, değerlendirme için temel gerçek işlevi gören, insanlar tarafından oluşturulan özetlerle ilişkilendirilir.TVSum veri kümesi aşağıdaki bileşenleri içerir:

**Videolar:** Veri kümesi, birkaç dakikadan birkaç saate kadar değişen uzunluklarda MP4 formatında video klipler içerir. Videolar, çok çeşitli konuları ve türleri kapsar ve video özetleme araştırması için çeşitli içerikler sağlar.

**İnsan tarafından oluşturulan özetler:** TVSum'daki her video, algoritmalar tarafından oluşturulan video özetlerinin kalitesini değerlendirmek için temel gerçek olarak kullanılan, insanlar tarafından oluşturulan birden çok özet ile ilişkilendirilir. Bu özetler, videoların ana içeriğinin kısa ve temsili açıklamalarını sağlar.

**Video meta verileri:** TVSum ayrıca her video için şov adı, bölüm başlığı, yayın tarihi ve video süresi gibi bilgiler dahil olmak üzere meta veriler sağlar. Bu meta veriler, bağlamsal analiz için veya videoları belirli kriterlere göre filtrelemek için kullanılabilir.

TVSum veri seti, video özetleme algoritmalarını değerlendirmek ve kıyaslamak için akademik araştırmalarda yaygın olarak kullanılır. Araştırmacıların F1 puanı, ROUGE puanı veya video geri çağırma gibi ortak değerlendirme ölçütlerini kullanarak farklı algoritmaların performansını karşılaştırmasına olanak tanıyan, insanlar tarafından oluşturulan özetlere sahip standartlaştırılmış ve farklı videolar sunar.

*D. Test Ortamı Ve Önerilen Yöntem*
  *D.1. Önerilen Yöntem*

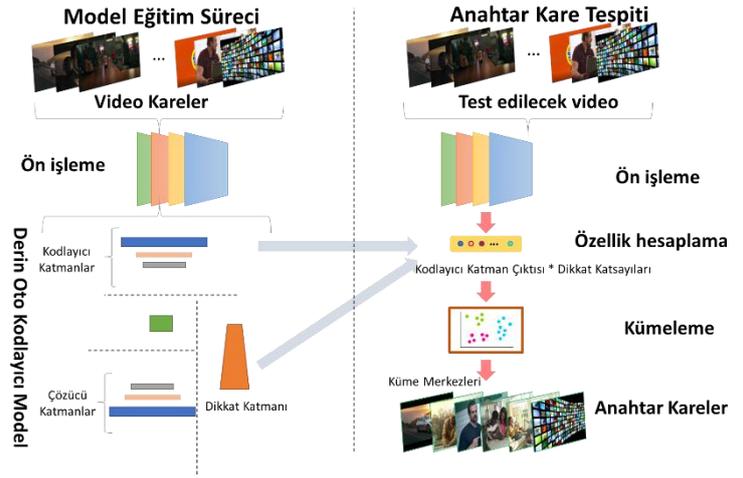

Şekil. 2. Model şeması

  *D.1.1.Model*

Çalışmada bir oto kodlayıcı kullanılmıştır, girdi verilerini tipik olarak daha düşük boyutlu bir gizli alana sıkıştırarak ve ardından orijinal biçimine geri döndürerek yeniden yapılandırmak üzere eğitilmiş bir tür sinir ağıdır.

Model mimarisi bir kodlayıcı ve bir kod çözücüden oluşur. Kodlayıcı, (3, 64, 64) şeklindeki giriş verilerini alır, düzleştirir ve ardından 'relu' aktivasyon fonksiyonlarıyla birkaç yoğun katmandan geçirir. Kodlayıcının çıktısı, gizli boyutun (64) boyutuna sahip bir vektördür. Kod çözücü daha sonra sıkıştırılmış gösterimi alır ve bir relu aktivasyon fonksiyonuna sahip yoğun katmanlar kullanarak orijinal giriş şekline geri döndürür.

Ayrıca modelin kodlayıcı kısmına dikkat katmanı eklenmiştir. Dikkat katmanı, kodlayıcının çıktısını alır, her zaman adımı için dikkat puanlarını hesaplar ve ardından bu puanları, dikkat ağırlıklı bir temsil oluşturmak için kodlayıcı çıktısına uygular. Ortaya çıkan dikkat ağırlıklı temsil daha sonra sabit boyutlu bir çıktı üretmek için küresel ortalama havuzlama katmanına beslenir.

Model daha sonra Adam iyileştirici ve ortalama kare hata kaybı işlevi kullanılarak derlenir ve doğrulanır.

  *D.1.2.Tespit Katmanı*

Bu adım, bir otomatik kodlayıcı ve kümeleme kullanan bir anahtar kare çıkarma algoritmasının bir uygulamasıdır. Algoritmanın amacı, videodaki en önemli anları ve olayları yakalayan bir video sekansından temsili kareler belirlemektir.

İlk olarak, kareleri HSV renk uzayına dönüştürerek, 64x64 piksel olarak yeniden boyutlandırarak ve piksel değerlerini 0 ila 1 aralığında normalleştirerek ön işler.

Ardından, özellikleri kümeler halinde gruplandırmak için KMeans kümeleme algoritmasını kullanır. Bu uygulamada küme sayısı farklı sayılara ayarlanabilir, ancak bu, belirli uygulamaya bağlı olarak ayarlanabilir.

Kümeleme tamamlandıktan sonra algoritma, özellik mesafesi açısından küme merkezine en yakın kareyi seçerek her küme için anahtar kareyi tanımlar. Bu anahtar kareler, orijinal dizideki indeksleri ve özellik gösterimleriyle birlikte saklanır.

Genel olarak, bu algoritma, dizideki diğer karelere benzerliklerine dayalı olarak bir video dizisinden anahtar kareleri ayıklamanın basit ama etkili bir yoludur. Ortaya çıkan anahtar kareler, videoyu özetlemek veya içeriğinin temsili bir görsel özetini sağlamak gibi çeşitli amaçlar için kullanılabilir.

Burada kullanılan kümeleme algoritmalarının küme sayısı parametresi aynı zamanda çıkarılacak olan anahtar kare sayısını da belirlemektedir. Algoritmanın sayı parametresine bağımlılığı olduğundan her videoda aynı miktarda kare olduğu var sayılacaktır. Bu durumda daha durağan videolarda anahtar kare sayısı daha az olacağından benzer karelerin tespiti ile karşılaşılacaktır. Veya daha yüksek sayıda tespit edilmesi gereken durumlarda daha az sonuç üretilmiş olacaktır. Bu durumların etkisini daha azaltmak için tespit edilen karelerin sayısı genelde kullanılandan daha yüksek belirlenip benzer karelerin temizlenmesi işlemi yapılması uygun görülmüştür. Bu işlem için bir uzaklık metriği ve uzaklıklar matrisi kullanılacaktır. Tespit edilen uzaklıklara göre birbirine yakın olan karelerden biri bulgu listesinden silinecek ve diğer temsili tutulacaktır. Yakınlık sınırını belirlerken istatistiksel güven aralıkları göz önünde bulundurulacak ve ortalama değerinden iki standart sapma daha düşük uzaklıkta olan kareler birleştirilecektir.

$$\mu = ortalama$$

$$\sigma = std.sapma$$

$$uzaklık < \mu - 2*\sigma \rightarrow birleştir \qquad (1)$$

*D.2. Ölçümleme Metodları*

Ölçümleme için kullanılan veri setinin anahtar kare düzeyinde bir etiket bilgisi olmadığından çıktının değerlendirmesinde uzman görüş kullanılacaktır. Anahtar kare olarak belirlenecek görüntünün video içerisinde yer alan her sahneyi tespit etmesi, eğer farklı sahneler yoksa görüntüdeki büyük değişiklikleri ve hareketleri tespit etmesi beklenecektir. Bu anlamda her video için anahtar kareler için kesit aralıkları belirlenecek ve test çıktısının bu aralıkta en az bir görüntü tespit edebilmiş olması beklenecektir. Elde edilecek veri ise her doğru tespit edilen ve tespit edilemeyen kareler için hata matrisi ile değerlendirilecektir.

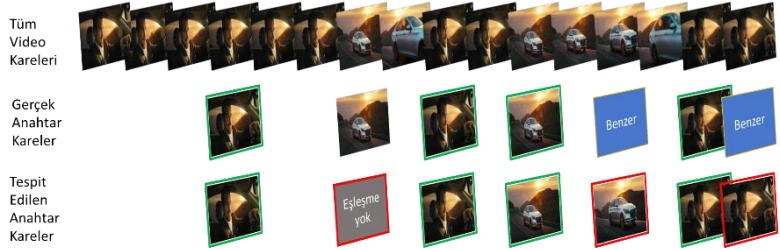

Şekil. 3. Tespit doğrulama durumları

Hata matrisi, bir makine öğrenimi modelinin performansını değerlendirmek için kullanılan bir tablodur. Bir ikili sınıflandırma problemi için gerçek pozitif (TP), gerçek negatif (TN), yanlış pozitif (FP) ve yanlış negatif (FN) değerlerinin bir matrisidir.

Bir ikili sınıflandırma probleminde, pozitif ve negatif olmak üzere iki olası sınıf vardır. Hata matrisi, tahmin edilen sınıf etiketlerini gerçek sınıf etiketleriyle karşılaştırır ve her sınıf için doğru ve yanlış tahminlerin sayısının bir dökümünü sağlar.

Matris tipik olarak dikey eksende gerçek sınıf etiketleri ve yatay eksende tahmin edilen sınıf etiketleri ile bir kare olarak temsil edilir. Matrisin dört değeri aşağıdakileri temsil eder:

Gerçek pozitifler (TP): aslında pozitif olan ve pozitif olarak doğru bir şekilde tahmin edilen örneklerin sayısı

Yanlış pozitifler (FP): aslında negatif olan ancak yanlışlıkla pozitif olarak tahmin edilen örneklerin sayısı

Yanlış negatifler (FN): aslında pozitif olan ancak yanlışlıkla negatif olarak tahmin edilen örneklerin sayısı

Gerçek negatifler (TN): gerçekte negatif olan ve doğru bir şekilde negatif olarak tahmin edilen örneklerin sayısı

Matristeki değerler, doğruluk, kesinlik, geri çağırma ve F1 puanı gibi bir makine öğrenimi modelinin performansını değerlendirmek için yaygın olarak kullanılan çeşitli ölçütleri hesaplamak için kullanılabilir. Bu metrikler, modelin ne kadar iyi performans gösterdiğine dair içgörüler sağlayabilir ve iyileştirme alanlarının belirlenmesine yardımcı olabilir.

### III. TEST

TABLO1: SINIF SONUÇLARI HATA MATRİSİ

|  |  | gerçek | |
|---|---|---|---|
|  |  | **0** | **1** |
| **tahmin** | **0** | 330 | 14 |
|  | **1** | 12 | 43 |

TABLO2: TEST SONUÇLARI BAŞARIM SKORLARI

| Sınıf | kesinlik | duyarlılık | f1-skor |
|---|---|---|---|
| **0** | 0,96 | 0,96 | 0,96 |
| **1** | 0,78 | 0,75 | 0,77 |

Kesinlik, modelin pozitif örnekleri doğru bir şekilde sınıflandırma yeteneğini gösterir. Bu durumda, sınıf 0 için kesinlik 0,96'dır, yani model pozitif sınıflandırılan tüm örneklerin %96'sını doğru bir şekilde sınıflandırdı. Ancak anahtar kare tespiti yöntemleri doğası gereği dengesiz sınıf problemleri olduğundan, normal sınıf için değerlendirme yapmak anlamsız olacaktır. Bu sebeple tespit edilen karelerin doğruluklarını incelemek gerekir. Sınıf 1 için kesinlik 0,78'dir, yani model pozitif olarak sınıflandırılan örneklerin %78'ini doğru bir şekilde sınıflandırdı. Duyarlılık, modelin pozitif örnekleri doğru bir şekilde tanımlama yeteneğini gösterir. Sınıf 1 için duyarlılık 0,75'tir, yani model sınıf 1'e ait tüm örneklerin %75'ini doğru bir şekilde tanımladı. F1 puanı, kesinlik ve duyarlılık arasındaki dengeyi gösteren bir ölçüttür. Bu durumda, sınıf 1 için F1 puanı 0,77'dir.

## IV. SONUÇ

Bu çalışmada, dikkat katmanına sahip derin bir otomatik kodlayıcı kullanan bir anahtar kare çıkarma yöntemi önerildi. Yöntem, önerilen yaklaşımın etkinliğini gösteren sınıflandırma görevinde 0,77'lik bir başarı oranı elde etti.

Yöntem, önce otomatik kodlayıcının kodlayıcı kısmını kullanarak video karelerinden özelliklerini çıkarır ve ardından K-means kümelemeyi kullanarak bu özellikleri kümeler. Anahtar kareler daha sonra, küme merkezine olan yakınlıklarına göre her bir kümeden seçilir. Otomatik kodlayıcının kodlayıcı kısmında dikkat katmanının kullanılması, video karelerindeki en belirgin özelliklerin vurgulanmasına yardımcı olur ve bu, anahtar kare çıkarma işleminin doğruluğunu artırır.

Deneysel sonuçlarımız, önerilen yöntemin, anahtar kare çıkarma için mevcut yöntemler kadar iyi performans gösterdiğini ve video özetleme ve eylem tanıma gibi çeşitli alanlarda potansiyel uygulamalara sahip olduğunu göstermektedir. Sınıflandırma görevindeki 0.77 başarı oranı, önerilen yöntemin oldukça doğru ve etkili olduğunu göstermektedir. Farklı modellerin simülasyon sonuçlarının araştırılması, bu makalede önerilen modelin iyi bir performans gösterebileceğini ortaya koymuştur.

Önerilen yöntemimiz, genellikle buluşsal yöntemlere veya videonun anlamsal içeriğini yakalayamayan düşük düzeyli özelliklere dayanan mevcut anahtar kare çıkarma yöntemlerinin sınırlamalarının dışına çıkar. Dikkat katmanına sahip derin bir otomatik kodlayıcı kullanarak, yöntemimiz video karelerindeki en göze çarpan bilgileri yakalayan üst düzey özellikleri ayıklayabilir. Ayrıca denetimsizdir, bu da onu ölçeklenebilir ve çok çeşitli video analizi görevlerine uygulanabilir kılar.

Gelecekteki çalışmalarımızda, ses ve hareket bilgileri gibi ek özellikleri dahil ederek yöntemimizin performansını daha da geliştirmenin yolları keşfedilecektir. Yöntemimizin etkinliğini spor videoları, haber videoları ve gözetleme videoları gibi farklı video türleri üzerinde de araştırması planlanmaktadır. Önerilen yöntemin video analizinde anahtar kare çıkarma için umut verici bir çözüm sağladığı ve video özetleme ve video alma(retrieval) gibi çeşitli uygulamalara uygulanabileceği görülmektedir.

Sonuç olarak, dikkat katmanına sahip derin bir otomatik kodlayıcı kullanan önerilen anahtar kare çıkarma yöntemi, video analizi için umut verici bir yaklaşımdır ve çok çeşitli alanlarda potansiyel uygulamalara sahiptir. Bu yöntemin aktif bir araştırma alanı olmaya devam edeceğine ve performansını ve doğruluğunu artırmak için daha fazla iyileştirme yapılabileceğine inanıyoruz.